\documentclass[lettersize,journal]{IEEEtran}
\usepackage{amsmath,amsfonts}
\usepackage{algorithmic}
\usepackage{algorithm}
\usepackage{array}
\usepackage[caption=false,font=normalsize,labelfont=sf,textfont=sf]{subfig}
\usepackage{textcomp}
\usepackage{stfloats}
\usepackage{url}
\usepackage{verbatim}
\usepackage{graphicx}
\usepackage{cite}
\usepackage{booktabs}
\usepackage[citecolor=blue, colorlinks]{hyperref}
\usepackage{xcolor}
\usepackage{soul}

\newcommand{\blfootnote}[1]{%
  {\let\thefootnote\relax\footnotetext{#1}}%
}

\begin{document}

\title{DecoRec: Decomposed 3D Scene Reconstruction from Single-View Images via Object-Level Diffusion}

\author{Yuhan Ping,
        Yuan Liu,
        Xiaoxiao Long,
        Peng Wang,
        Junhui Hou,
        Jianyi Zheng,
        Jia Pan,
        Xin Li,
        Cheng Lin$^\dagger$
        % <-this % stops a space
\thanks{Yuhan Ping, Yuan Liu, Xiaoxiao Long, Peng Wang and Jia Pan are with the Department of Computer Science, the University of Hong Kong. Emails: \{csyhping, yuanly, xxlong, totoro97\}@connect.hku.hk, jpan@cs.hku.hk} \thanks{Junhui Hou is with the Department of Computer Science, City University of Hong Kong. Email: jh.hou@cityu.edu.hk} 
\thanks{Jianyi Zheng is with the Faculty of Humanities and Arts, Macau University of Science and Technology. Email: jyzheng@must.edu.mo} 
\thanks{Xin Li 
is with the Department of Computer Science and Engineering, Texas A\&M University. Email:  xinli@tamu.edu
}\thanks{Cheng Lin is with the Department of Computer Science and Engineering, Macau University of Science and Technology. Email: chlin@connect.hku.hk}% <-this % stops a space
%\thanks{Manuscript received April 19, 2021; revised August 16, 2021.}
}

\markboth{Manuscript submitted to IEEE TVCG}%
{Shell \MakeLowercase{\textit{et al.}}: A Sample Article Using IEEEtran.cls for IEEE Journals}

\IEEEpubid{0000--0000/00\$00.00~\copyright~2021 IEEE}

\maketitle

% \begin{abstract}
\begin{abstract}

In this paper,  we introduce \textit{DecoRec}, a novel system designed to elevate single-view 2D images to a decomposed 3D scene mesh. Current methods for single-view scene reconstruction typically rely on object retrieval or the regression of coarse 3D voxels or surfaces, leading to inaccuracies in capturing the appearance and geometry of the input image. The lack of high-quality large-scale scene-level datasets further complicates direct 3D scene generation from single-view images.
To achieve high-quality 3D scene generation from a single-view image, DecoRec takes advantage of recent diffusion-based single-view object reconstruction methods to reconstruct individual objects separately. 
Subsequently, a refinement pipeline is proposed to effectively merge these reconstructed objects, enhancing appearance and geometry through a differentiable rendering technique and diffusion-guided refinement.
Our results demonstrate that DecoRec facilitates high-quality single-view scene reconstruction in both geometry and novel synthesis, offering significant benefits for downstream applications like room interior design.
\end{abstract}

\blfootnote{$\dagger$ Corresponding author}

% \end{abstract}


\section{Introduction}

\IEEEPARstart{R}{econstructing} 3D indoor scenes is a crucial task for room interior design. Typically, a design process begins with creating 2D designs, which are then transformed into 3D models through 3D modeling. A finely crafted 3D model not only offers superior visual representation but also facilitates design analysis. Recent advancements in 2D image diffusion models~\cite{rombach2022high} have significantly aided designers by automatically generating 2D designs from simple text prompts, allowing them to concentrate on creative tasks without the burden of tedious drawing. However, converting these 2D designs into precise 3D models remains a challenging and time-consuming task.

\begin{figure}[htb]
  \includegraphics[width=0.47\textwidth]{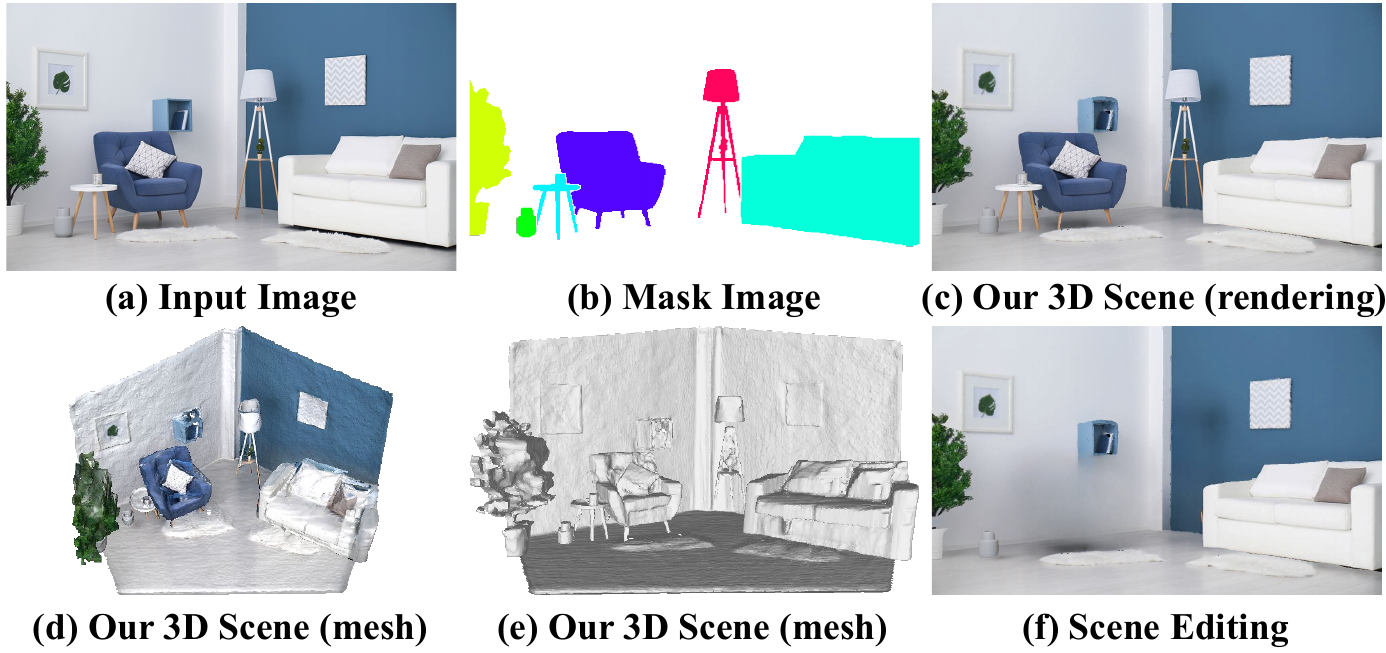}
  \caption{Given a single viewpoint photo of an indoor scene or an image of an interior space design draft (a) and the corresponding object masks (b), our method DecoRec generates the corresponding 3D scene (c-e), which allows explicit control of each 3D object for scene editing, as shown in (f). 
  }
  \label{fig:teaser}
\end{figure}

Many computer graphics or vision researchers have tried to automate this single-view 3D reconstruction problem, but the 3D reconstruction quality is still far from satisfactory. Some works~\cite{cao2022monoscene,popov2020corenet,nie2020total3dunderstanding} address this problem by regressing semantic voxels or surfaces from single-view images by neural networks, which can produce completed scene meshes but with low quality in both geometry and appearances. Other works~\cite{yan2023psdr,gumeli2022roca,kuo2021patch2cad,wu2023generalizing} take advantage of large-scale 3D model datasets as a database to retrieve the most similar objects in the database. These retrieval-based methods enable decomposed and high-quality 3D reconstruction of the scene if the objects are contained in the database. However, when there are no exactly matched objects in the database, these methods fail to reconstruct these objects with fidelity. Recent advancements~\cite{poole2022dreamfusion,liu2023syncdreamer,liu2023zero,shi2023zero123++,long2023wonder3d} on diffusion models for single-view reconstruction have enabled high-quality reconstruction of single objects. However, these methods are trained on the object-level dataset Objaverse~\cite{deitke2023objaverse} and are not applicable to scene reconstruction due to the lack of 3D scene-level data. Thus, how to accurately reconstruct 3D scenes from single-view images is still an open problem.
\IEEEpubidadjcol

In this paper, we present a new system called DecoRec to lift a single-view image into a decomposed 3D scene. To address the problem of reconstructing a complex indoor scene from just a single-view image, DecoRec essentially consists of a Segmentation-Reconstruction-Refinement (SRR) pipeline. Given an indoor scene image, the image can be captured from real world, rendered from synthetic data, or designed by stylized artists, we first apply the image segmentation method~\cite{kirillov2023segment} to segment the input image into several object masks and a background region. Then, for each segmented object mask, we run a single-view object reconstruction algorithm~\cite{wang2024crm} to reconstruct a coarse mesh from the given single-view image. Finally, in the refinement stage, we adjust the poses, appearances, and geometry of all coarse object meshes by inverse rendering~\cite{Laine2020diffrast} and diffusion-based refinement~\cite{brooks2023instructpix2pix}. This allows us to reconstruct a set of decomposed high-quality 3D meshes for the whole scene. As shown in Fig.~\ref{fig:teaser}, our reconstruction results not only facilitate free-viewpoint rendering in the reconstructed scene but also can be easily edited by replacing or removing objects.

\begin{figure}
    \centering
    \includegraphics[width=0.47\textwidth]{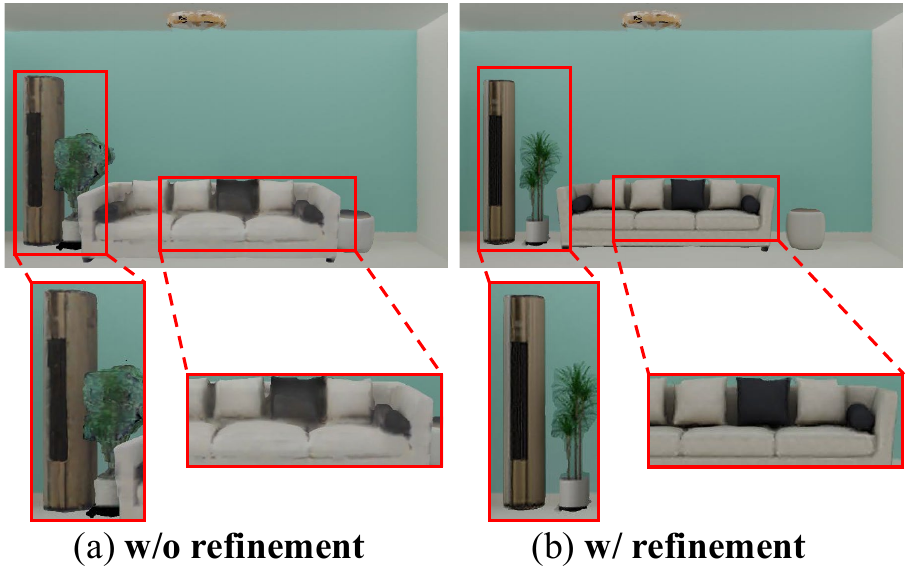}
    \caption{Refinement comparison. Single-view reconstruction method cannot provide accurate results. With our refinement, our method can achieve high-quality scene reconstruction.}
    \label{fig:refine-compare}
\end{figure}

Essentially, we find that simply reconstructing each object from the object masks using single-view reconstruction methods does not lead to accurate and high-quality reconstruction, as shown in Fig.~\ref{fig:refine-compare}. Our refinement significantly increases the fidelity of the reconstructed meshes with the input image and the visual quality of novel viewpoints. Specifically, in our refinement stage, we rely on a single-view depth estimation to roughly estimate the location of each object and place the reconstructed object meshes in their corresponding location. Then, we apply the differentiable mesh rendering to further refine the poses, appearances, and geometry of these coarse meshes. To further improve the visual quality of novel viewpoints deviating from the input image, we adopt the diffusion model InstructPix2Pix~\cite{brooks2023instructpix2pix}. The InstructPix2Pix model gradually biases the rendered novel-view images to the appearances of the input image to improve the visual quality.

We conduct experiments on both generated 2D interior room designs from text-to-image models and images from the Internet. All results show that our method outperforms existing single-view scene reconstruction methods by a large margin in terms of visual quality and fidelity. We further demonstrate applications like editing the reconstructed 3D scenes to help designers improve their efficiency in 2D interior design.

In summary, our major contributions are listed as follows.
\begin{itemize}
    \item We propose a novel pipeline that lifts a 3D decomposed scene from a single RGB image. We integrate and utilize several pre-training modules and propose refinement strategies to achieve the overall task objectives.
    \item We develop a two-stage coarse-to-refine framework to improve both the geometry quality of 3D objects in the scene and the rendering quality of novel view synthesis.
    \item Qualitative and quantitative experiments show that our method outperforms existing methods on both real-world and synthetic scene data.
\end{itemize}

\section{Related Work}

Scene reconstruction is a well-studied task in the computer vision community. Previous works mainly concentrate on reconstructing a 3D scene from multiview images~\cite{schoenberger2016mvs} or depth sensors~\cite{izadi2011kinectfusion}. Our paper only requires a single-view image as input to reconstruct the whole scene. 

\subsection{Single-view Scene Reconstruction}

Single-view 3D scene reconstruction is an ill-posed problem, which strongly relies on the priors~\cite{ho2020denoising} to reconstruct reasonable 3D geometry. Early-stage works~\cite{mazur2023superprimitive,nie2020total3dunderstanding,dahnert2021panoptic,cao2022monoscene,popov2020corenet,li2024know,yao2023depthssc,yao2023ndc,cao2023scenerf,li2023rico} mainly directly regress the geometry of the scene from the images. These methods represent the scene with occupancy voxels~\cite{cao2022monoscene,popov2020corenet,li2024know,yao2023depthssc,yao2023ndc}, surfaces~\cite{nie2020total3dunderstanding,dahnert2021panoptic,li2023rico}, primitives~\cite{mazur2023superprimitive} and Radiance Field~\cite{cao2023scenerf}. Recent works~\cite{zhang2021holistic,liu2022towards,chen2024single} have similar pipelines as \cite{nie2020total3dunderstanding} but focus on improving the object reconstruction quality, but they are still limited and fail to recognize and reconstruct unseeded objects in the scene.
Many of these works~\cite{cao2022monoscene,popov2020corenet,yao2023depthssc,yao2023ndc,chu2023buol,zhang2023uni} focus on the semantic-related completion task. These regression-based methods often have difficulty in generalizing to unseen data and the geometry reconstruction is usually not very accurate but just shows the spatial locations of objects. To get a more accurate geometry reconstruction and better textures, another branch of works~\cite{gumeli2022roca,kuo2021patch2cad,wu2023generalizing,chu2024open,gao2023diffcad,yan2023psdr} resorts to object retrieval to find the most similar object in a large-scale database. Such retrieval-based reconstruction methods are able to get a detailed high-quality 3D reconstruction of the objects in the scene. However, when there are some unseen scene objects in the object database, these methods often fail to correctly reconstruct the scene with fidelity. 
In comparison with these works, our method relies on a diffusion-based single-view object reconstruction method~\cite{wang2024crm} to show strong generalization ability and reconstruct high-quality textured meshes. Two concurrent works~\cite{dogaru2024generalizable,chen2024comboverse} also adopt the diffusion-based single-view reconstruction method. \cite{dogaru2024generalizable} adopts a similar pipeline as ours but our method has an essential refinement stage to improve the visual quality and the consistency with the input image. \cite{chen2024comboverse} adopts a new distillation-based single-view reconstruction algorithm to generate multiple objects while our method does not require distillation but feeds the image into a single-view reconstruction to directly get 3D meshes. 

\subsection{Novel-view Synthesis from Single-view}  Some works~\cite{li20233d,bao2023sine,wang2023lolep,wang2023perf,zhang2023deformable,pu2023sinmpi,han2022single,chung2023luciddreamer, gao2024cat3d, shriram2024realmdreamer, mirzaei2023spin,szymanowicz2025bolt3d} also implicitly reconstruct a scene from a single-view image for novel-view-synthesis(NVS). These methods are mainly based on the prior from the single-view depth estimation~\cite{han2022single} or diffusion models~\cite{chung2023luciddreamer}. \cite{gao2024cat3d} trains a multi-view diffusion model from a large-scale multiview dataset which can perform mutl-view sampling from the input image and achieve novel view synthesis based on NeRF\cite{mildenhall2021nerf} scene representation, \cite{szymanowicz2025bolt3d} implements a similar idea but uses 3D Gaussian Splatting(3DGS)\cite{kerbl20233d} as the scene representation. \cite{shriram2024realmdreamer} extends novel views by inpainting RGB and depth image pairs to reconstruct a 3DGS scene representation. \cite{mirzaei2023spin} also achieves novel view synthesis but it aims at multi-view segmentation and perceptual inpainting. Novel-view-synthesis does not require reconstructing the scene explicitly into 3D surfaces, while our method not only enables NVS but also reconstructs decomposed 3D meshes for the scene.

\subsection{3D Scene Generation}

Our method can be regarded as scene generation from a single-view image, which also relates to the 3D scene generation methods. Some works~\cite{tang2024diffuscene,zhai2024commonscenes,lin2024instructscene,zhang2024furniscene,zhai2024echoscene,wu2024blockfusion,zhang2023scenewiz3d,gao2023graphdreamer,hwang2023text2scene} concentrate on generating a layout or a scene graph from a text prompt or other user inputs and then generating decomposed 3D meshes for each component. To generate a scene, another branch of works~\cite{zhang2024text2nerf,zhou2024dreamscene360,li2024dreamscene,ouyang2023text2immersion,schult2023controlroom3d,fang2023ctrl,zhang20243d,mao2023showroom3d,ma2024fastscene,wang2024360dvd} first generate an image or a panorama and then reconstruct the scene from the generated image or panorama. Almost all of these methods are mainly designed for the text-to-scene generation task and these methods mainly focus on the first step to generate a layout or a high-quality panorama. Instead, our method mainly concentrates on generating a decomposed 3D scene from an image, which can be incorporated with a text-to-image model for scene generation.

\subsection{Single-view Object Reconstruction} 

Recent developments in diffusion models~\cite{poole2022dreamfusion,shi2023mvdream,long2023wonder3d,liu2023syncdreamer,liu2023zero,shi2023zero123++,wang2023imagedream,lu2023direct2,jun2023shape,zou2023triplane,li2023instant,tang2023dreamgaussian} have enabled high-quality 3D object reconstruction from single-view images. Most of these methods are trained on the Objaverse~\cite{deitke2023objaverse,deitke2024objaverseXL} dataset. The main trend of these methods is to utilize the multiview diffusion~\cite{tang2023MVDiffusion,liu2023syncdreamer} to generate multiview consistent images and then reconstruct the 3D models from the generated multiview consistent images. However, directly applying such object reconstruction methods to single-view scene reconstruction is not feasible, which produces unrealistic and incorrect reconstruction results. Our method is based on these object reconstruction methods and tackles each object separately.
\begin{figure*}[t]
  \includegraphics[width=\textwidth]{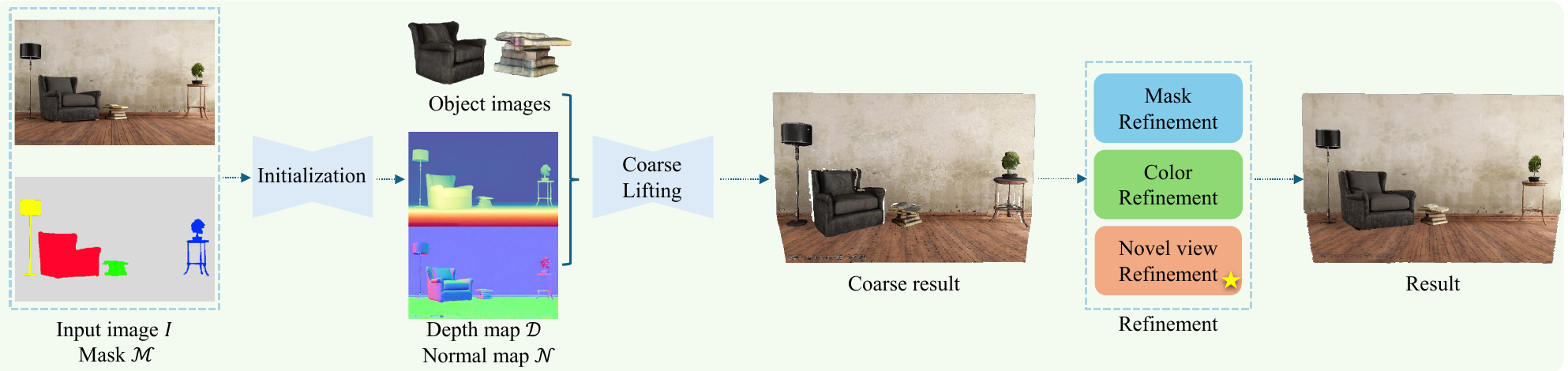}
  \caption{Overview of our pipeline. Given an input image and object masks, our method first performs a coarse decomposition and reconstruction for both objects and background. Then, we refine the foreground objects and background, including mask, color, and novel view refinement, to achieve high-quality 3D scene reconstruction for both consistent novel view synthesis and high-quality 3D geometry.}
  \label{fig:overview}
\end{figure*}

\section{Proposed Method}

Fig.~\ref{fig:overview} provides an outline of our approach named DecoRec. We begin by taking the input image associated with object masks to generate depth and normal maps. 
Then, we utilize a single-view estimation technique to derive coarse meshes for individual objects and perform background inpainting for a fully realized background mesh. 
Finally, we refine the coarse object meshes and the background meshes to get 
an improved overall reconstruction outcome. In what follows, we will detail each module.

\subsection{Initialization}
\label{sec:single}

Let $I$ denote the input image, and the associated object masks can be divided into a background mask $\mathcal{M}_{b}$ and a collection of foreground masks $\{\mathcal{M}_{f}^{(i)} | i=1,...,N\}$ encompassing all $N$ objects. Initially, we utilize a single-view depth and normal estimation technique called GeoWizard ~\cite{fu2024geowizard}, denoted as $G$, to derive a depth map $\mathcal{D}$ normalized in $[0,~1]$ and a normal map $\mathcal{N}$ from  $I$. Note that the estimated depth map $\mathcal{D}$ is affine-invariant with unknown scale $s$ and shift $t$, we have to recover a metric depth map $\mathcal{\hat{D}} = \mathcal{D} \times s + t$ from it. We first compute a normal map $\mathcal{\hat{N}}$ from the estimated depth map $\mathcal{D}$ by a least square fitting operation, then minimize the difference between $\mathcal{\hat{N}}$ and $\mathcal{N}$ to optimize $s$ and $t$.

\subsection{Coarse Lifting}
\label{sec:recon}
In this stage, we aim to reconstruct both the foreground objects and backgrounds from the images, the object masks, and the depth map. 

\subsubsection{\textbf{Coarse object reconstruction}} Taking an object mask, we first crop the region of the object and mask out the background to get a single foreground object $I_f^{(i)}$. Then, we feed $I_f^{(i)}$ to a single-view 3D reconstruction method called Convolution Reconstruction Model (CRM)~\cite{wang2024crm}. CRM first generates multiview images and so-called canonical coordinate maps (CCMs) from the $I_f^{(i)}$. Then, a CNN is applied to regress a triplane-based SDF representation from the multiview images and CCMs. A colored mesh $F_i$ can be extracted from the triplane SDF. The reconstructed mesh is coarsely aligned with the input image. For each object mask, we repeat this CRM-based reconstruction pipeline to reconstruct every object in this scene. Though CRM produces a reasonable reconstruction for these objects, the coarse meshes often show blurry textures and unaligned geometry. We will refine the geometry, appearances, and poses of this coarse reconstruction in our refinement.

\subsubsection{\textbf{Background reconstruction}} 
\begin{figure}[t]
    \centering
    \includegraphics[width=0.47\textwidth]{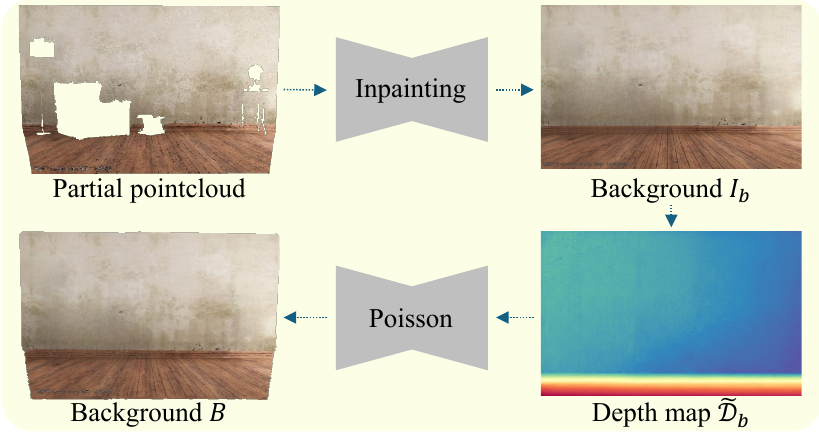}
    \caption{Background reconstruction. We reconstruct a complete background mesh from the mask-removed partial point cloud.}
    \label{fig:bgrecon}
\end{figure}

As shown in Fig.~\ref{fig:bgrecon}, by removing all the depth of the foreground objects in $\mathcal{D}$, we can get a partial point cloud of the scene. 
To make it more complete, we first apply an inpainting network called LaMa~\cite{suvorov2021resolution} to fill the holes of these foreground objects to get a complete background image $I_b$. Then, we again apply the GeoWizard to estimate a background depth map $\mathcal{\hat{D}}_b$ again and align the scale of this background depth map with the partial foreground depth map. Finally, a Poisson surface reconstruction algorithm is applied to the background depth points to extract a colored mesh $B$ for the background of this scene. 

\subsection{Refinement}
\label{sec:refine}

Given the reconstructed coarse meshes for the background and the foreground objects, 
we perform a joint refinement of their geometry, spatial positioning, orientation, and surface appearance. 
This is accomplished through a dual-optimization strategy. First, we employ differentiable mesh rendering to enforce pixel-level consistency between the reconstructed scene and the original input image. Second, we introduce a diffusion-based novel-view refinement stage; this process enhances the visual fidelity of synthesized perspectives while ensuring they remain semantically and structurally consistent with the source observation.

\subsubsection{\textbf{Differentiable rendering}}
\label{}
For every coarse object mesh $F_i$, we associate a trainable affine transformation $\mathcal{A}_i$ with it to transform it to a more accurate shape and location $\mathcal{A}_i \circ F_i$.
We utilize the differentiable rendering framework NvDiffRast~\cite{Laine2020diffrast} $\eta$ to render all the foreground and background meshes by
\begin{equation}
    \{\hat{M}_f^{(i)}\},\hat{D},\hat{I}=\eta(\{\mathcal{A}_i \circ F_i\},B, \pi_{in}),
\end{equation}
where $\{\hat{M}_f^{(i)}\}$ are the set of rendered masks for all foreground objects, $\hat{D}$ is the rendered depth map, $\hat{I}$ is the rendered images from all the foreground and background meshes, and $\pi_{in}$ means the camera pose for the input image. Then, we compute the following losses to optimize the appearances and geometry of all meshes as well as all the object poses.

\begin{equation}
\begin{split}
\begin{aligned}
\ell_\text{inv} &=\ell_\text{mask}+\ell_\text{rgb}+\ell_\text{depth}, ~~
\ell_\text{mask} =\sum_i \|\hat{M}_i - M_i\|^2_2,\\
\ell_\text{rgb} &=\|\hat{I}-I\|_2^2,~~\ell_\text{depth} =\|\hat{D}-D\|^2_2.
\end{aligned}
\end{split}
\end{equation}

To minimize the above loss, the trainable parameters here include the vertex-wise colors, vertex locations of all these meshes, and also the affine transformations $\mathcal{A}_i$. The mask loss $\ell_\text{mask}$ and depth loss $\ell_\text{depth}$ act like anchors to regularize the locations and size of each object. Without a depth loss, the object can incorrectly shrink its size and move closer to the camera but still satisfy the mask and rendered colors. The color loss $\ell_\text{rgb}$ constrains the colors of these meshes to be consistent with the input image.

\subsubsection{\textbf{Diffusion-based novel-view refinement}}

We further improve the rendering quality of novel-view synthesis from these foreground and background meshes. We achieve this novel-view refinement with the diffusion prior from the diffusion model InstructPix2Pix~\cite{brooks2023instructpix2pix}. InstructPix2Pix takes a text prompt $y$ and a condition image $I_c$ as input and generates an image, which is a modified version of $I_c$ according to the text prompt $y$. In our pipeline, we do not aim to revise the image but want to improve the visual quality of novel-view images and their consistency with the input image. 

\begin{figure}[t]
    \centering
    \includegraphics[width=0.47\textwidth]{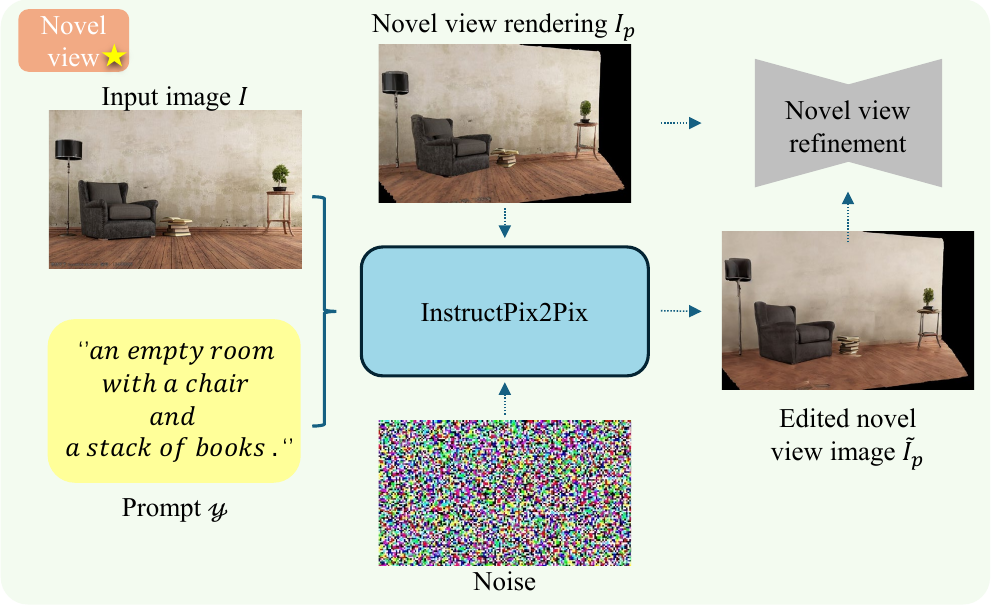}
    \caption{Illustration of the diffusion-based novel-view refinement. We utilize a diffusion model to improve both the visual quality and consistency with the input image of novel-view rendering images.}
    \label{fig:dfnvrefine}
\end{figure}

Details of the novel-view refinement are shown in Fig.~\ref{fig:dfnvrefine}. Specifically, given a rendered novel-view $I_p$ images from meshes $I_p=\eta(\{\mathcal{A}_i \circ F_i\},B,\pi_{new})$ on a new viewpoint $\pi_{new}$ using our differentiable renderer $\eta$ on the pose, we will use the InstructPix2Pix model to refine it by a diffusion-guided loss.
For the InstructPix2Pix model, we take the input image $I$ as the conditional image $I_c=I$. Then, some text prompts are generated on the input image $I$ by OpenCLIP~\cite{ilharco_2021_5143773}, which serves as the text prompt $y$. Subsequently, we do not generate images from scratch because this would simply lead to the same generation results as $I$. Instead, we follow \cite{haque2023instruct} to add $t$-level noise to the rendered image $I_p$ and then feed the noisy version to the InstructPix2Pix model to regenerate a noise-free image $\tilde{I}_p$. Finally, we compute the diffusion loss by
\begin{equation}
    \ell_\text{diff}=\|I_p - \tilde{I}_p\|^2,
\end{equation}
We apply $\ell_\text{diff}$ to optimize the meshes' appearances to improve the visual quality of rendered novel images.

\section{Experiments}
\subsection{Experiment Settings}

\subsubsection{\textbf{Implementation details}}
Our work is mainly implemented with Pytorch{~\cite{paszke2019pytorch}} on an Nvidia GeForce RTX 4090 GPU. In the subsequent description of the implementation details, we use corresponding official default parameters, except those declared on purpose. The segmentation masks are provided by SegDrawer\footnote{https://github.com/lujiazho/SegDrawer} built on SAM \cite{kirillov2023segment}. The masked object images are preprocessed to be centered with a grey background ($R\text{:}127, G\text{:}127, B\text{:}127$). We use CRM{~\cite{wang2024crm}} to generate the 3D model for segmented objects with the preprocessed masked images. For background reconstruction, we utilize the LaMa~\cite{suvorov2021resolution} in-painting network, we use the model named $big\text{-}lama$ and we set the dilate kernel size as $30$ for better results. We employ GeoWizard~\cite{fu2024geowizard} for depth estimation; we set the ensemble size to $3$ and the domain to $indoor$. We use NvDiffRast~\cite{Laine2020diffrast} for differential rendering, and we use its $cuda$ initialization. Novel view editing is performed by InstructPix2Pix~\cite{brooks2023instructpix2pix}, we use a diffusion step of $10$, and we follow~\cite{haque2023instruct} to add $t$-level noise to the rendered image. The text prompt for describing the scene is generated by OpenCLIP~\cite{ilharco_2021_5143773}, we use the model named $coca\_ViT\text{-}L\text{-}14$ with pretrained keys $mscoco\_finetuned\_laion2B\text{-}s13B\text{-}b90k$. For the optimization procedure, we use the Adam optimizer{\cite{kingma2014adam}} with a learning rate $1e^{-3}$. There are approximately $7500$ iterations for the optimization procedure, and we use an edit rate of every $10$ iterations for novel view optimization. The camera pose for the input image is set to the origin and facing the z-axis. The focal length is estimated from the diagonal length of the image. For novel viewpoints in the refinement stage, we implement a spiral trajectory like NeRF\cite{mildenhall2021nerf} by sampling novel camera poses that differ from the original camera pose in three dimensions, $0.4\times$ scene range on the x-axis, $0.4\times$ scene range on the y-axis and $0.16\times$ scene range on the z-axis.

\subsubsection{\textbf{Datasets}}
We test our method on both real-world scenes and synthetic scenes. Real-world scene data are mainly acquired from ShutterStock\footnote{https://www.shutterstock.com/} and Google Images, while synthetic scene data are extracted from 3D-Front~\cite{fu20213d} and 3D-Future~\cite{fu20213d2} which contain high-quality indoor scenes and decomposed furniture models. Due to the texture missing issue for walls and floors in the original datasets, we also use some textures from ambientCG\footnote{https://ambientcg.com/}. 8 real-world scenes and 11 synthetic scenes are used in the evaluation. To demonstrate the generalizability of our method, we also test our method on stylized scenes. 8 stylized scenes are collected from Dribbble\footnote{https://dribbble.com/}. We include all our test cases and corresponding results in the qualitative comparison and result gallery in later sections.

\subsubsection{\textbf{Metrics}} 
For quantitative measurement, we report several image and geometry quality metrics widely used in novel view synthesis problems, including (1) PSNR, (2) SSIM, (3) LPIPS~\cite{zhang2018perceptual}, (4) Chamfer Distance, and (5) F-Score. For PSNR and SSIM, we utilize the implementation in scikit-image\cite{van2014scikit}, and for LPIPS, we use its official implementation. For F-Score, we use a 
$CD < 0.1$ threshold. For CD, we measure each reconstructed object and report the average results. For real-world scenes, we evaluate the CLIP Score\cite{taited2023CLIPScore} by comparing the input view between input images and reconstructed scenes.

\subsubsection{\textbf{Baseline methods}}

We compare our method with several baseline methods, including
Poisson~\cite{kazhdan2006poisson}, CRM~\cite{wang2024crm}, AdaMPI~\cite{han2022single}, ROCA~\cite{gumeli2022roca}, Total3D\cite{nie2020total3dunderstanding}, InstPIFu\cite{liu2022towards}, SSR\cite{chen2024single}, Uni3D\cite{zhang2023uni} and Gen3DSR\cite{dogaru2024generalizable}. Poisson is a traditional 3D reconstruction method that reconstructs the mesh from the point cloud which is extracted from the depth map used in our pipeline. CRM utilizes generative models to perform image-to-3D conversion, and we use the input scene images as the input to CRM. AdaMPI achieves view synthesis from single-view images with adaptive multiplane images, which do not reconstruct 3D meshes. Thus, we mainly compare the novel view synthesis quality with AdaMPI. ROCA is a retrieval-based method for single-view scene reconstruction. We use the official codes and datasets of ROCA for retrieval. Total3D, InstPIFu, and SSR all detect and reconstruct 3D objects from single-view images and predict their poses. Uni3D integrates panoptic segmentation from single-view images and reconstructs the scene with semantics. Gen3DSR is a concurrent method that is similar to our coarse stage.

\subsection{Comparisons with Baselines} 
\begin{figure*}[htp] 
    \centering
    \includegraphics[width=\textwidth]{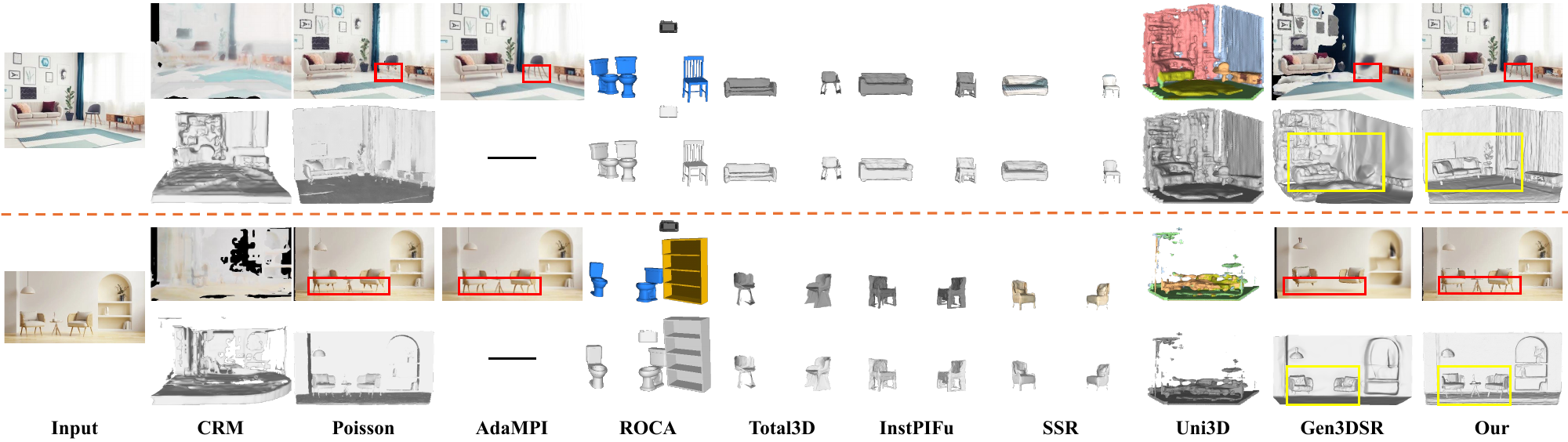}
    \caption{Qualitative comparison with other methods on the real-world dataset. From top to bottom, we show one novel-view rendering and the corresponding 3D geometries. Our method outperforms in both view synthesis and 3D geometry qualities. Some obvious structure distortions are marked in red and geometry differences are marked in yellow.}
    \label{fig:comp_baseline}
\end{figure*}

\begin{figure*}[htp]
    \centering
    \includegraphics[width=\textwidth]{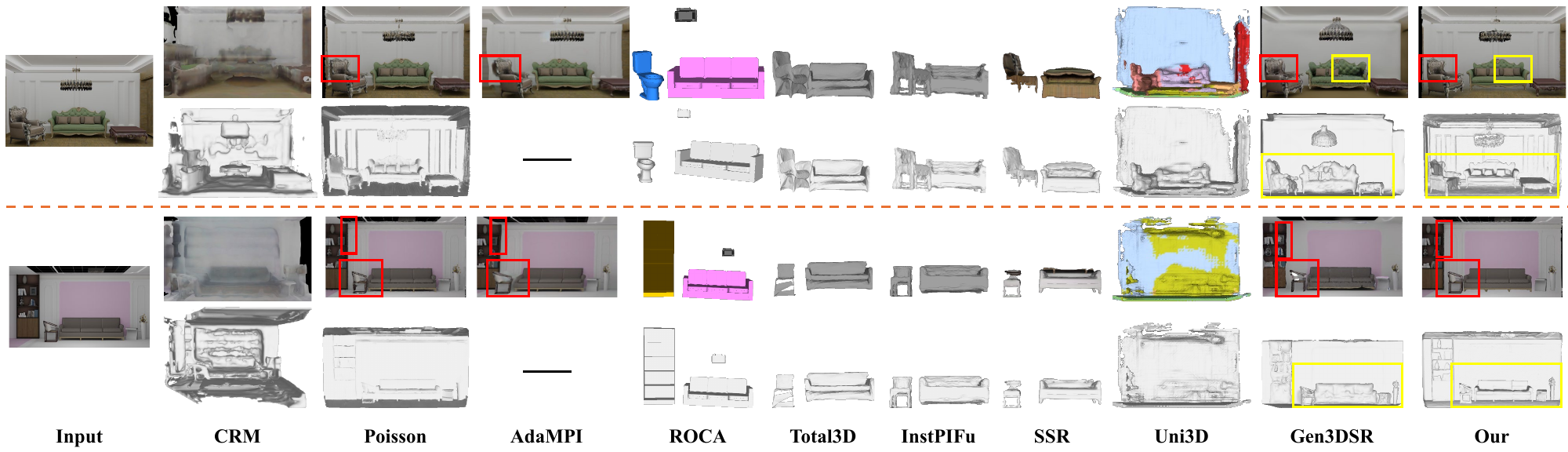}
    \caption{Qualitative comparison with other methods on the synthetic dataset. From top to bottom, we show two novel views and the corresponding 3D geometries. Our method outperforms in both view synthesis and 3D geometry qualities. Some obvious structure distortions are marked in red and geometry differences are marked in yellow.
    }
    \label{fig:comp_baseline2}
\end{figure*}

\begin{figure*}[!h]
    \centering
    \includegraphics[width=\textwidth, height=4.25in]{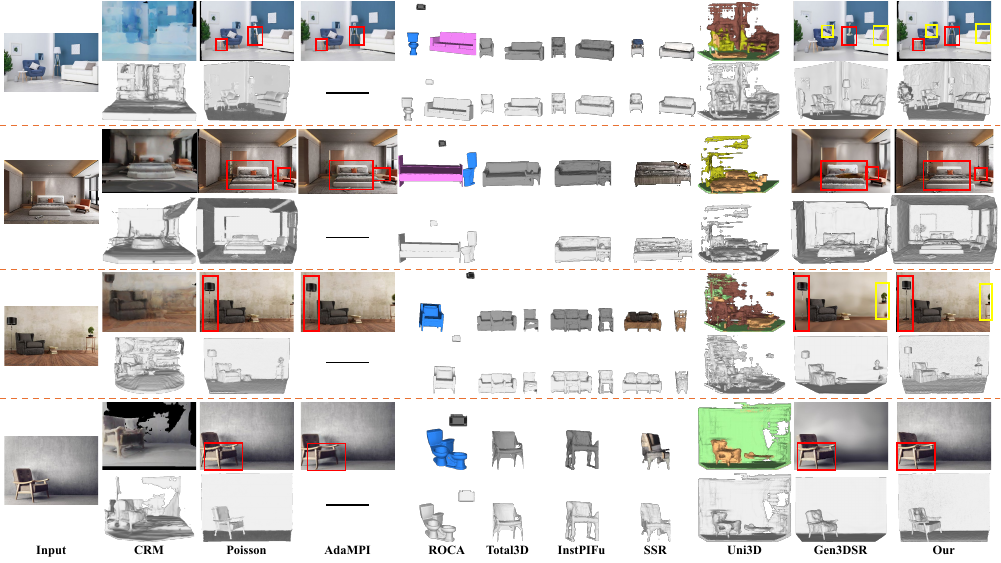}
    \caption{Result gallery of qualitative comparison with other methods. 
    }
    \label{fig:res_gal_1}
\end{figure*}

\begin{figure*}[!t]
    \centering
    \includegraphics[width=\textwidth, height=4.25in]{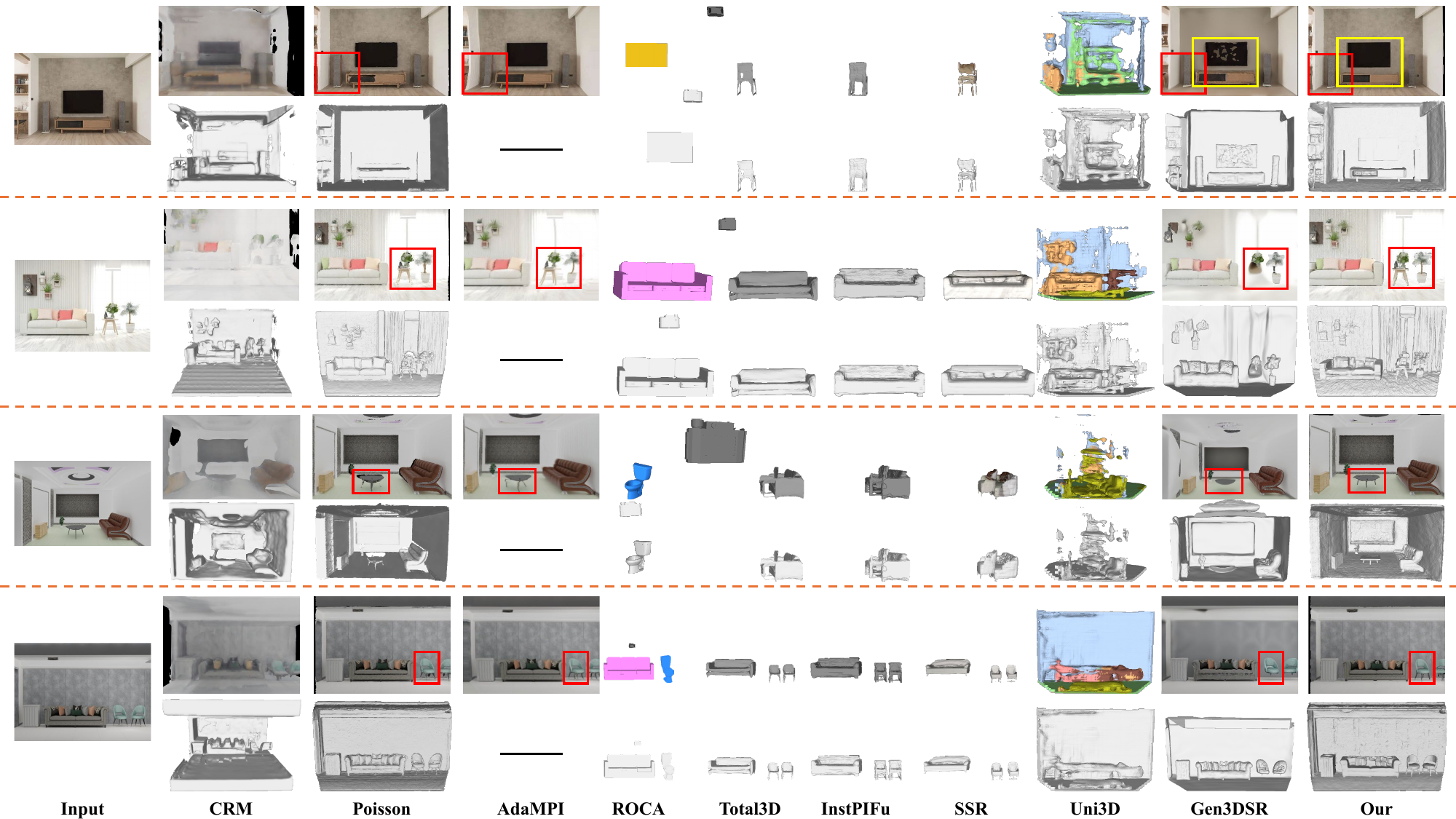}
    \caption{Result gallery of qualitative comparison with other methods. 
    }
    \label{fig:res_gal_2}
\end{figure*}

The qualitative comparisons of real-world and synthetic scenes are shown in Fig.~\ref{fig:comp_baseline} and Fig.~\ref{fig:comp_baseline2}. We also provide a result gallery with all scenes that we tested and corresponding results, as shown in Fig.~\ref{fig:res_gal_1}, Fig.~\ref{fig:res_gal_2} and Fig.~\ref{fig:res_gal2_1}. The quantitative comparisons are shown separately since some baselines are not designed for novel view synthesis or object reconstruction, as shown in Tab.~\ref{tab:metric} and Tab.~\ref{tab:metric2}. 
CRM utilizes generative models to convert 2D images to 3D meshes. However, since it is only trained on the object datasets, CRM cannot produce reasonable results for the single-view scene reconstruction. Poisson is a traditional reconstruction method that conducts meshes from estimated single-view depth. Though it is able to reconstruct the input view with high fidelity, it cannot guess the unseen parts and strongly relies on the quality of the monocular depth map, producing distorted surfaces as shown in the figure. 
AdaMPI achieves novel view synthesis from MPIs but still suffers from distortion on novel viewpoints, especially for regions with thin structures. 
Besides, without a 3D mesh as its geometry, AdaMPI has no abilities or flexibility to perform scene editing. 
ROCA retrieves CAD models for each object in the dataset, and the retrieval results rely highly on the datasets. If a similar object does not exist in the database, the resulting objects will be incorrect in both semantics and appearances. Total3D, InstPIFu, and SSR 
heavily rely on the 2D detection results. If an object is out of the category of the 2D detector, these methods fail in 3D reconstruction. 
The results of Uni3D contain severe distortion if the semantic segmentation is not accurate. Gen3DSR is a concurrent compositional generation method that adopts a similar pipeline as our coarse stage. However, our method contains an essential refinement stage to improve the visual quality, showing more consistency with the input and geometry details than Gen3DSR. 
Our method clearly outperforms all baselines by reconstructing plausible and accurate scenes from the given input image. Our method can generate high-quality 3D geometry, enabling the rendering of consistent novel-view images and advanced scene editing operations. 
 
\begin{table}[!t]
\caption{Quantitative novel view synthesis results on the synthetic dataset. Our method produces better rendering quality, as indicated by PSNR, SSIM, and LPIPS, than all baseline methods. CLIP Score is also provided for the real-world dataset.
}
\centering
\begin{tabular}{lrrrr}
\toprule
        & PSNR$\uparrow$ & SSIM$\uparrow$ & LPIPS$\downarrow$ & CLIP-S$\uparrow$ \\
\midrule
Poisson & 19.3205  & 0.6571      & 0.2188  & 0.9627 \\
CRM     & 13.9334  & 0.6149      & 0.5270  & 0.6425 \\
AdaMPI  & 21.9959  & 0.7489      & 0.1852  & \textbf{0.9850} \\
Gen3DSR & 20.1648  & 0.7277      & 0.2056  & 0.8502 \\
Our     & \textbf{23.2017}  &\textbf{0.7919} & \textbf{0.1502} & 0.9724\\   
\bottomrule
\end{tabular}

\label{tab:metric}
\end{table}

\begin{table}[htb]
\caption{Quantitative object reconstruction results on the synthetic dataset. Our method produces better reconstruction quality, CD and F-Score, than all baseline methods.
}
\centering
\begin{tabular}{lrrrr}
\toprule
        & CD-O$\downarrow$ & F-Score$\uparrow$ \\ 
\midrule
Poisson     & 0.1281     & 45.1023 \\
CRM         & 0.1708     & 24.3121 \\
ROCA        & 0.1551     & 29.9654 \\
Total3D     & 0.1802     & 34.6879 \\
InstPIFu    & 0.1479     & 39.2105 \\
Uni3D       & 0.4295     & 16.8127  \\
SSR         & 0.1368     & 39.4872  \\
Gen3DSR     & 0.1349     & 53.7818  \\
Our         &\textbf{0.1071}  & \textbf{63.2835}\\   
\bottomrule
\end{tabular}

\label{tab:metric2}
\end{table}

\subsection{Lifting Stylized Scenes}
\begin{figure}[hbp] 
    \centering
    \includegraphics[width=0.47\textwidth]{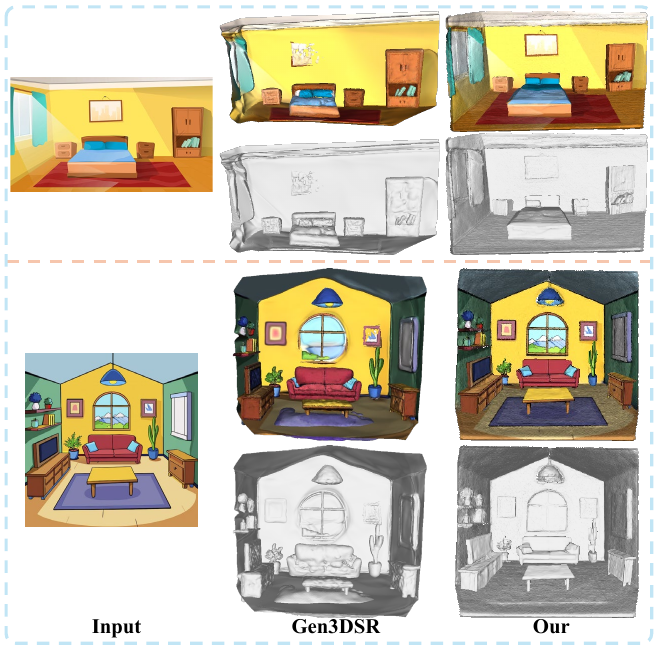}
    \caption{Qualitative comparison with Gen3DSR on stylized inputs.}
    \label{fig:supp_comp_cartoon_1}
\end{figure}

To fully demonstrate the generalization ability of our method, we also collect some stylized scenes by designers from Dribble. Since other baselines struggled to provide reasonable results given such special input, we only compare our method with Gen3DSR~\cite{dogaru2024generalizable}. Qualitative results are shown in Fig.~\ref{fig:supp_comp_cartoon_1}. 
Our method generates scenes with finer details and textures on both objects and backgrounds. The reconstructed objects' details are obviously sharper and more accurate than Gen3DSR. The performance indicates our method has a strong generalization capability.

\subsection{Ablation Study}
\subsubsection{\textbf{Loss functions}}
To fully demonstrate the design of our method, we conducted ablation studies based on different combinations of losses we mentioned in the Method section. We denote ``Full model" as our work with all losses. ``w/o mask loss" means without the mask loss $\ell_\text{mask}$. ``w/o color loss" means without the color loss $\ell_\text{rgb}$. ``w/o diff loss" means without the diffusion loss $\ell_\text{diff}$. 

\begin{figure*}[htp]
    \centering
    \includegraphics[width=\textwidth]{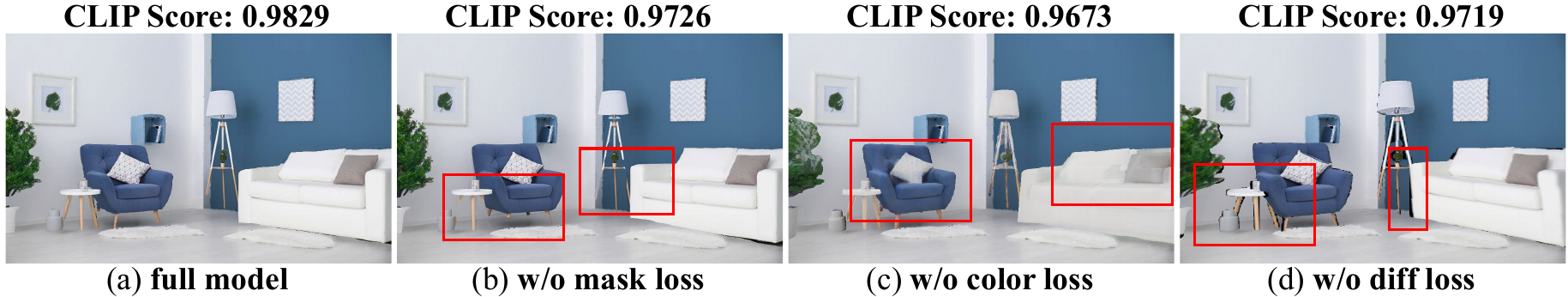}
    \caption{Ablation studies on different loss settings. Without any designed losses, it will cause issues like texture blurring, structure distortion, and black borders.}
    \label{fig:ablation}
\end{figure*}

The ablation results are shown in Fig.~\ref{fig:ablation}. In ``w/o mask loss" (Fig.~\ref{fig:ablation}(b)), the object positions are not aligned precisely with the corresponding mask on the input image. 
Similarly, without color loss (Fig.~\ref{fig:ablation}(c)), the position of objects can be optimized, but the colors of the surfaces are not consistent with the input images. Moreover, there is an overall blurring of textures. In ``w/o diff loss" (Fig.~\ref{fig:ablation}(d)), there are obvious black silhouettes around objects, and the reconstruction results are not accurate. Simply optimizing the input images leads to deteriorated results on novel view renderings, which can only be resolved by refining the novel view images. 
Our full method provides accurate and plausible scene reconstruction with all necessary designed losses, as shown in Fig.~\ref{fig:ablation}(a).

\begin{figure*}[htp]
    \centering
    \includegraphics[width=\textwidth]{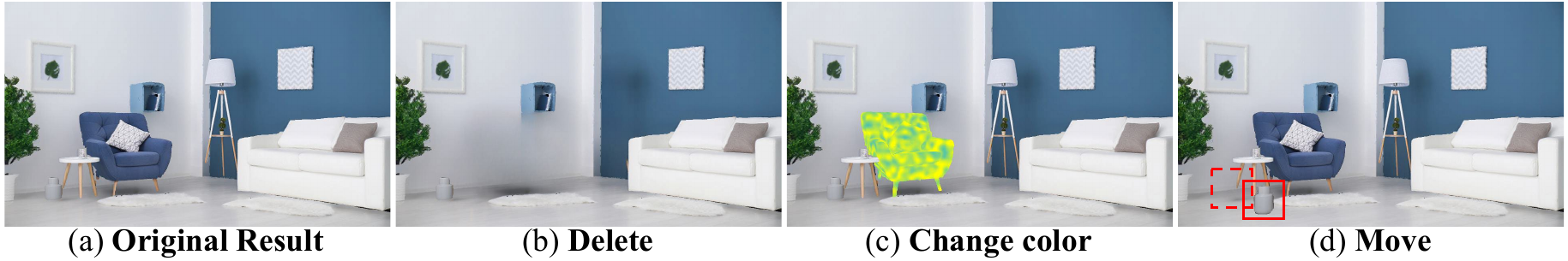}
    \vspace{-15pt}
    \caption{Flexibility of scene editing. Our method enables decomposed 3D lifting of 2D images and thus, users can perform different editing operations including deleting, color changing, and moving.}
    \label{fig:edit}
\end{figure*}

\subsubsection{\textbf{Change modules}}
\label{sec:change}
\begin{figure}[!h] 
    \centering
    \includegraphics[width=0.47\textwidth]{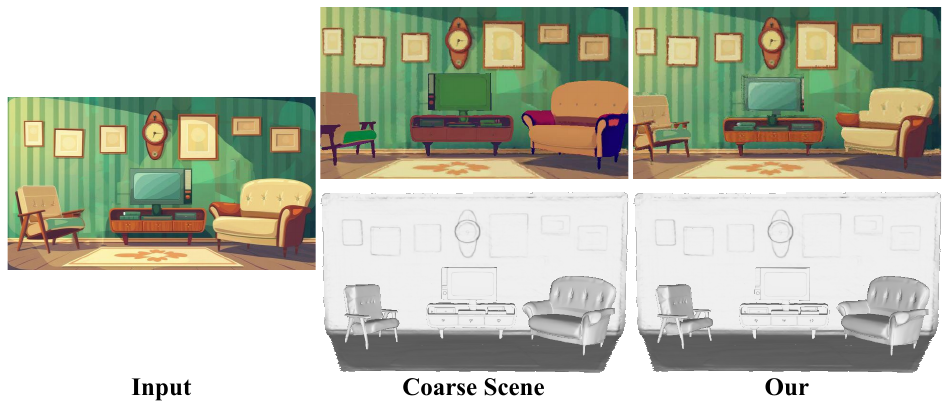}
    \vspace{-10pt}
    \caption{Additional results with implementation of Hunyuan3D-2 and DepthPro. }
    \label{fig:h3d}
\end{figure}

Our method has no specific requirement on the object reconstruction and depth estimation models. We chose CRM and GeoWizard because these methods show great power in corresponding tasks. Our method has a flexibility that supports replacing modules like the object reconstruction and depth estimation models. We have added additional results using other SOTA methods at this stage. For the object reconstruction model, we use Hunyuan3D-2\cite{hunyuan3d22025tencent,yang2024hunyuan3d, lai2025hunyuan3d25highfidelity3d}. For the depth estimation model, we use DepthPro\cite{Bochkovskii2024:arxiv}. The result is shown in subsection G of the Experiment section, Fig.~\ref{fig:h3d}. By changing to newly developed models, the coarse result's quality is improved, but they still deviate from the input image(the shape, color, position and orientation of objects), so a refinement stage proposed by our method still makes a great improvement to get a better result.

\subsubsection{\textbf{Coarse and refinement stages}}
\begin{figure}[!h] 
    \centering
    \includegraphics[width=0.47\textwidth]{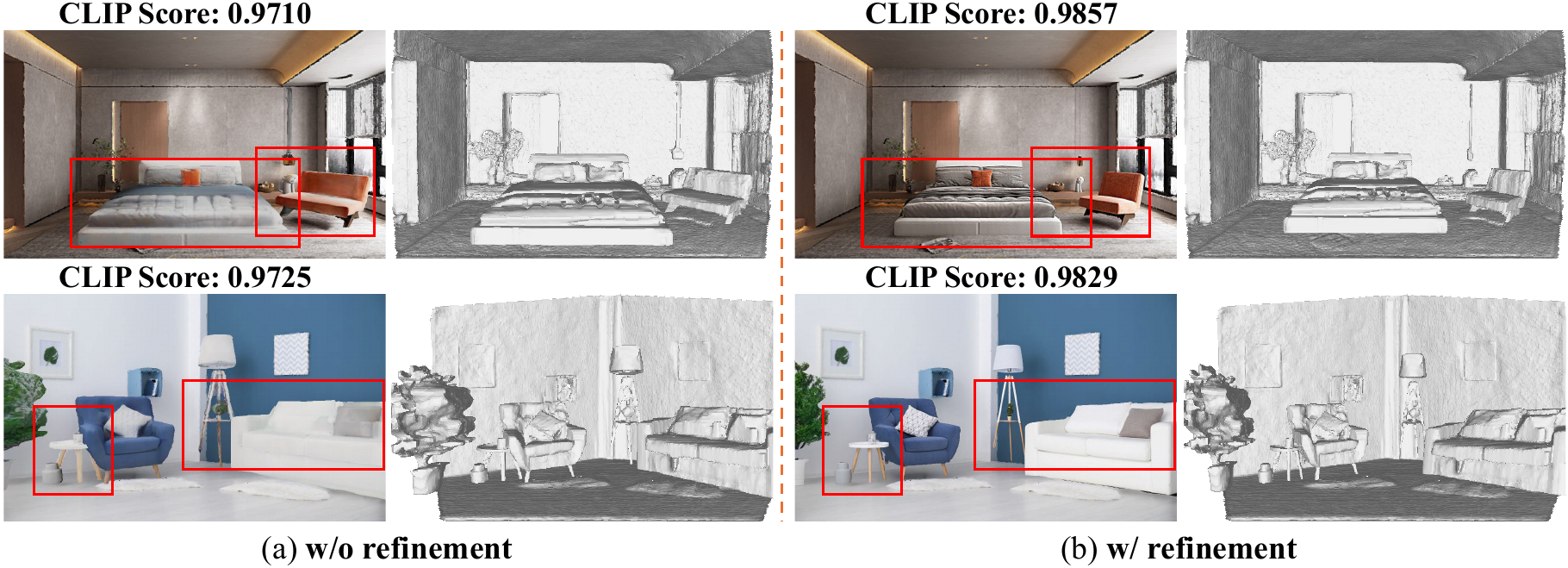}
    \vspace{-10pt}
    \caption{More comparisons between coarse and refinement stages. }
    \label{fig:coarse_refine}
\end{figure}

The proposed refinement stage is a significant design in our method. To better demonstrate the effectiveness and novelty of the refinement stage, we present additional results comparing the coarse and refinement stages, as shown in Fig.~\ref{fig:coarse_refine}. It is obvious that the coarse stage can provide a roughly aligned scene, but the shape, position and color of objects are not accurate. With our proposed refinement stage, we can optimize a better 3D scene with correct geometry, location and appearance. When changing the object reconstruction and depth estimation modules with more advanced models like Hunyuan3D-2\cite{hunyuan3d22025tencent, yang2024hunyuan3d, lai2025hunyuan3d25highfidelity3d} and Depthpro\cite{Bochkovskii2024:arxiv}, the quality of the coarse scene improves but still differs from the input image. The improvement introduced by the refinement stage is still significant, as shown in Fig.~\ref{fig:h3d}. 

\subsection{Editing the Scene}

As discussed above, our method can lift a 2D image into a decomposed 3D scene, which means we have both high-quality renderings and the corresponding 3D geometries. Consequently, unlike those methods that perform view synthesis only, such as AdaMPI, we have the 3D meshes as the backbone so that our method provides the ability to achieve scene editing, such as modifying colors and replacing or removing objects from the scene. An editing example is shown in Fig.~\ref{fig:edit}. From Fig.~\ref{fig:edit}(b) to (d), we demonstrate object deletion, color modification, and object movement, which shows a high flexibility of scene editing.

\begin{figure*}[ht]
    \centering
    \includegraphics[width=\textwidth]{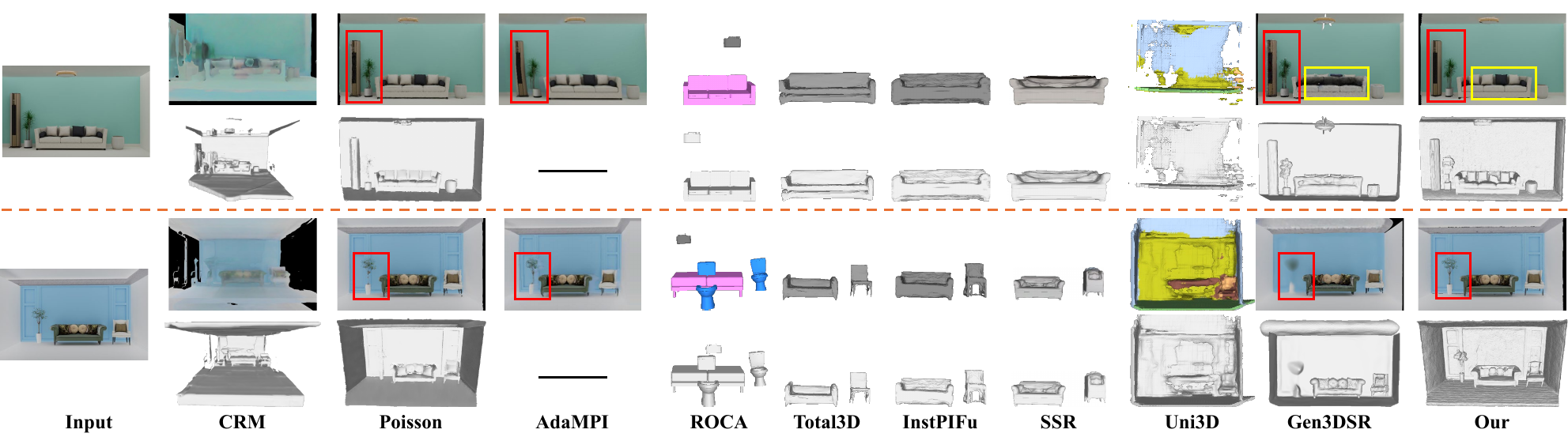}
    \vspace{-15pt}
    \caption{Result gallery of qualitative comparison with other methods.}
    \label{fig:res_gal2_1}
\end{figure*}

\subsection{More Analysis}
\subsubsection{\textbf{More scene types}}
\begin{figure}[!h] 
    \centering
    \includegraphics[width=0.47\textwidth]{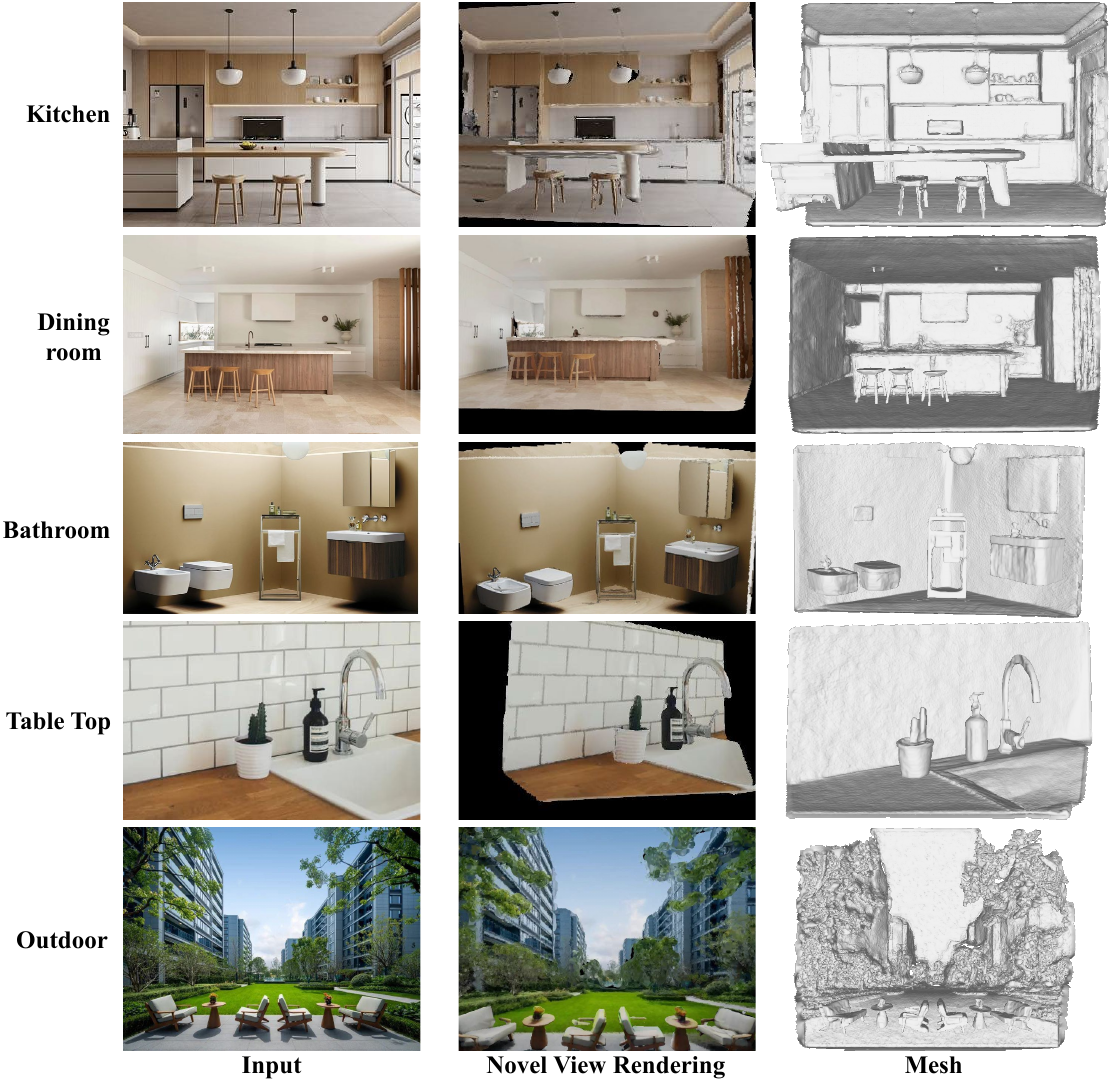}
    \vspace{-10pt}
    \caption{Result of our method on more types of scenes, including kitchen, dining room, bathroom, table top and even door scenes.}
    \label{fig:more_type}
\end{figure}

Our proposed method can generate plausible 3D scenes for a wide range of input images, including real-world, synthetic and stylized scenes. To fully demonstrate the generality of our proposed method, we conduct additional experiments on more types of scenes as shown in Fig.~\ref{fig:more_type}. We show more scene types like kitchen, dining room, and bathroom. Besides, our method is not inherently limited to indoor scenes, we also show table top and even outdoor scenes. Furthermore, we have included comprehensive qualitative comparisons in the earlier sections, and we present some extreme angles for these results in the supplementary materials. While we do not anticipate significant deviations in the camera perspective during optimization, this is mainly because the extreme viewpoints are not represented in the pretrained models. Results show that our method can generate plausible 3D scenes across a wide range of scene types and still has good quality from extreme camera angles. Our current goal is to primarily lift indoor scenes from a single image. In future work, we will expand our method to accommodate a more general concept of scenes. 

\subsubsection{\textbf{More comparisons with Gen3DSR}}
\begin{figure}[!h] 
    \centering
    \includegraphics[width=0.47\textwidth]{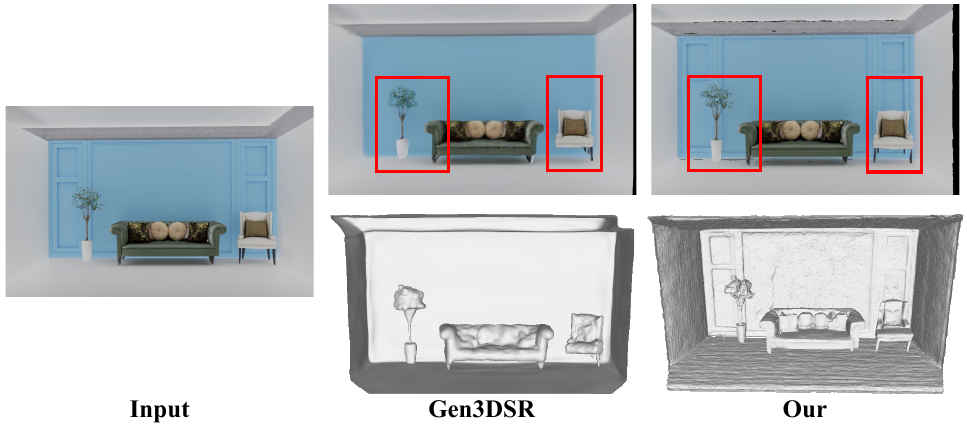}
    \vspace{-10pt}
    \caption{Comparison with Gen3DSR by giving the same mask as our method.}
    \label{fig:same_mask}
\end{figure}

To enable a more in-depth comparison, we conduct additional comparisons with Gen3DSR on multiple dimensions. Firstly, we give Gen3DSR the same mask image as our method because the default segmentation method in Gen3DSR sometimes cannot provide correct masks for all foreground objects (i.e., second row in Fig.~\ref{fig:res_gal2_1}, the plant on the left is not segmented by Gen3DSR). The result is shown in Fig.~\ref{fig:same_mask}. By using the same mask as our method, Gen3DSR can segment the plant on the left correctly. However, it is obvious that the mesh quality of our method outperforms Gen3DSR with more geometric details and better shapes.

\begin{figure}[!h] 
    \centering
    \includegraphics[width=0.47\textwidth]{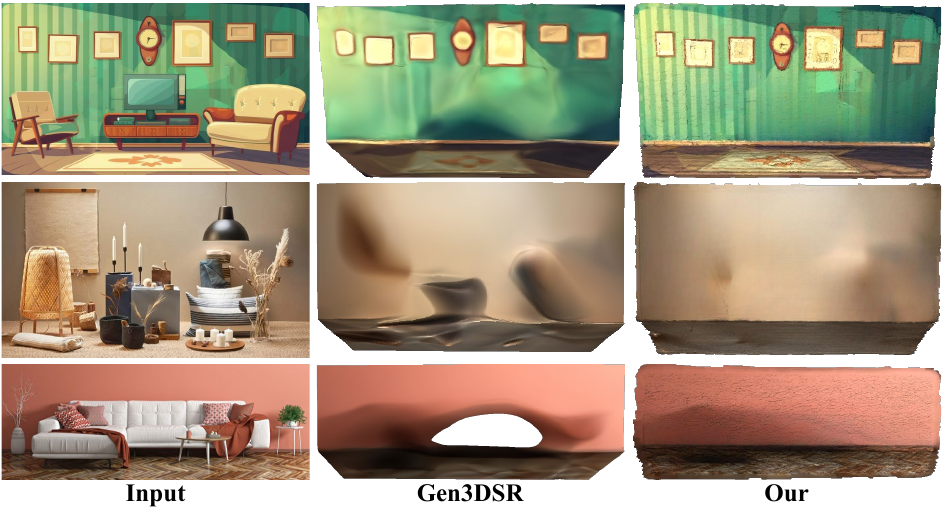}
    \vspace{-10pt}
    \caption{Comparisons of background reconstruction results. The middle and bottom rows show scenes with many foreground objects and only a little area of the background is visible. }
    \label{fig:bg_comp}
\end{figure}

Secondly, we also show the background reconstruction quality between Gen3DSR and our method because the background is also essential in 3D scenes as shown in Fig.~\ref{fig:bg_comp}. We give Gen3DSR the same mask as our method. Gen3DSR reconstructs the background by implementing an MLP to extract the SDF of the background and expects the network to interpolate the missing areas. If the scene contains many objects and only a little area of the background is visible, it fails to cover the hole region and cannot interpolate correct colors as shown in Fig.~\ref{fig:bg_comp} middle and bottom rows. While our method leverages an inpainting model to fill the holes, the occluded area can be completed with reasonable and plausible content to get a satisfying background. Besides, even in normal scenes like the top row in Fig.~\ref{fig:bg_comp}, our method can provide a background with more details that align with the input.

\subsection{Inference Time}
Our method is implemented on a single Nvidia GeForce RTX 4090 GPU. For an input image with resolution 1200*800, the background inpainting by LaMa takes about 3 seconds. The mesh reconstruction stage takes about 43 seconds for each object using CRM, about 1-3 minutes using Hunyuan3D (specifically, about 34 seconds for shape generation, about 35-200 seconds for texture generation). The mesh reconstruction time increases linearly with the number of segmentation results, similar to existing compositional methods like Gen3DSR. The mesh reconstruction procedure can be further accelerated by implementing parallelization if more GPU resources are available. The background reconstruction takes about 25 seconds if using GeoWizard for depth estimation, about 55 seconds if using DepthPro (the major difference is that GeoWizard predicts a normal map but DepthPro doesn't, we have to compute the normal map from DepthPro's depth map using Open3D). The refinement stage takes about 5-10 minutes, depending on the complexity of the scene. We will work on improving the time efficiency in our future work.

\subsection{Limitations}

Despite its effectiveness, our approach has certain limitations that provide avenues for future research. Currently, the pipeline does not explicitly model surface materials or environmental illumination. Consequently, it lacks the ability to render accurate view-dependent effects, such as specular highlights or physically consistent shadows, especially during scene editing. Integrating material synthesis and lighting estimation from single-view images would likely resolve these artifacts. Furthermore, while the 3D reconstruction module partially mitigates object-to-object occlusions, some mask shapes remain incomplete, please refer to the supplementary for details. We intend to address this in future work by incorporating diffusion-based 2D or 3D inpainting techniques to achieve more robust occlusion handling.

\section{Conclusion}

In this paper, we introduced a novel framework, namely DecoRec, for lifting decomposed 3D scenes from single-view indoor images. DecoRec integrates individual object reconstruction with background depth estimation and inpainting to establish a coarse 3D representation. To achieve high-fidelity results, we employed a multi-stage refinement process that optimizes object placement and appearance through a combination of mask, color, and diffusion-based losses. Specifically, we leveraged the InstructPix2Pix model within our diffusion loss to enhance rendering quality across novel viewpoints. Comprehensive evaluations on both synthetic and real-world datasets demonstrate the superior effectiveness and robustness of our method.

\section*{Acknowledgments}
This research is partially supported by the Natural Science Foundation of China under Grant 62422118 and the Hong Kong Research Grants Council under Grants 11219324 and N\_CityU1114/2.

\bibliographystyle{IEEEtran}
\bibliography{refs}

% \end{thebibliography}

\end{document}